%% file: main.tex
\documentclass{article}

\usepackage[final]{corl_2026} % Uncomment for the camera-ready ``final'' version.

\usepackage{amsmath} 
\usepackage{xcolor}
\usepackage[most]{tcolorbox} 
\usepackage[nolist,nohyperlinks]{acronym}
\usepackage{graphicx}
\usepackage{amssymb}
\usepackage{amsfonts}
\usepackage{cite}
\usepackage{authblk}
\usepackage{multirow}
\usepackage{booktabs}
\usepackage{subcaption}
\usepackage{float}
\usepackage{wrapfig}
\input{acronym}

\usepackage{xstring}
\title{
Spiking Neural Network Control of a Flapping-Wing Robot on Resource-Constrained Hardware
}

\author[1,2]{Rim El Filali}
\author[1]{Chenrui Feng}
\author[1,3,$\dagger$]{Chao Gao}
\author[1,$\dagger$]{Weibin Gu}
\affil[1]{Institute for AI Industry Research (AIR), Tsinghua University, Beijing, PR China}
\affil[2]{Department of Computer Science and Technology, Tsinghua University, Beijing, PR China}
\affil[3]{Xinchen Qihang Inc., Beijing, PR China}
\affil[$\dagger$]{Correspondence: \texttt{chao.gao@cantab.net}, \texttt{guweibin@air.tsinghua.edu.cn}}

\begin{document}
\maketitle

%===============================================================================

\begin{abstract}
Flapping-Wing Micro Aerial Vehicles (FWMAVs) provide exceptional maneuverability and aerodynamic efficiency but pose significant challenges for onboard control due to nonlinear dynamics and stringent Size, Weight, and Power (SWaP) constraints, as exemplified by a butterfly-inspired robot less than 30~gram.
%a 26-gram butterfly-inspired robot. #TODO: Replace this in the camera-ready paper
To this end, we present a hierarchical neuromorphic control framework that enables fully onboard, closed-loop flight on a widely available, resource-constrained ESP32 microcontroller with a unit cost of approximately \$5. Specifically, our method deploys two lightweight Spiking Neural Networks (SNNs) onboard: one for state estimation from raw sensory feedback and another for control via modulation of a Central Pattern Generator (CPG) for wing actuation. Trained by imitation learning, the system achieves stable pitch and heading angle tracking during untethered real-world flight. Experimental results further reveal that the SNN-based controller reduces latency by 36\% (1059~$\mu s$ to 680~$\mu s$) and power by 18\% (0.033~W to 0.027~W) for inference compared to the conventional Artificial Neural Network (ANN) baseline, demonstrating the viability of spike-based computation without specialized hardware. To the best of our knowledge, this work constitutes the first demonstration of fully onboard neuromorphic control for autonomous flight of a FWMAV, highlighting the potential of SNNs to enable energy-efficient autonomy under stringent SWaP constraints. 

\noindent
\textbf{Visual abstract:} \href{https://tinyurl.com/airpulse-snn-imitation}{\texttt{tinyurl.com/airpulse-snn-imitation}}\\[0.25em]
\textbf{Code:} \href{https://github.com/NEAR-the-future/Espikify.git}{\texttt{github.com/NEAR-the-future/Espikify}}

% \textbf{Visual abstract}: \url{https://tinyurl.com/airpulse-snn-imitation} 

% \textbf{Code}: \url{https://github.com/NEAR-the-future/Espikify.git}
\end{abstract}


\section{Introduction}
\label{sec:introduction}

Over the last decade, micro aerial vehicles have become increasingly prevalent thanks to their maneuverability, affordability, and relative ease of control. Among these, \acp{FWMAV} represent a promising class of bio-inspired robots that emulate the flight mechanics of flapping insects and vertebrate flyers~\citep{phan2019insect, nekoo2025survey, hao2025insect}. Considerable progress has been made in their development with platforms such as RoboBee~\citep{jafferis2019untethered}, DelFly~\citep{de2009design}, KUBeetle~\citep{phan2017design}, Purdue Hummingbird~\citep{zhang2017design}, Bat Bot~\citep{ramezani2017biomimetic}, and LisRaptor~\citep{phan2024twist}, to list a few. Unlike fixed- or rotary-wing vehicles, flapping-wing systems can remain aerodynamically efficient at low Reynolds numbers by leveraging unsteady lift mechanisms. Their lightweight structures and compliant wings also make them inherently collision-tolerant, enabling non-invasive operation in confined and cluttered environments without posing risks to humans or the surroundings~\citep{nekoo2025survey}. This opens up a wide range of applications such as crop surveying and chemical plume detection, where traditional drones usually fall short.

Yet achieving robust autonomous flight with \acp{FWMAV} remains challenging. Their highly nonlinear, coupled, and time-varying aerodynamics become even harder to handle at small scales due to structural compliance and fabrication variability. As constructing a reliable reduced-order model for controller design is particularly difficult under these conditions, recent work has turned to learning-based approaches. Paradigms such as \ac{RL} and \ac{IL} offer a way to bypass explicit modeling by optimizing wing kinematics and learning control policies directly from data~\citep{xiong2023lift, cai2025learning, sharvit2025deep, hsiao2025rtmpc_il}. However, most existing studies rely on conventional \acp{ANN}, whose dense computations incur high latency and energy costs on \ac{SWaP}-constrained microcontroller units, limiting real-time onboard control.

To this end, neuromorphic computing offers a promising pathway toward lightweight, energy-efficient embedded autonomy~\citep{bing2018survey}. Typically, \acp{SNN} encode information through sparse spike events and operate in an event-driven manner, thereby reducing inference cost on resource-constrained hardware~\citep{eshraghian2023training, oikonomou2025reinforcement}. However, their application to closed-loop robotic control remains limited, with few real-world demonstrations in \ac{UAV} systems. Early research on SNN-based control of insect-scale flapping-wing robots was restricted to simulation~\citep{clawson2016spiking, clawson2017adaptive}. Recent work demonstrated neuromorphic estimation and control pipelines on quadrotors operating at high control rates on embedded hardware~\citep{stroobants2025neuromorphic}, and fully spiking \ac{RL} policies trained with \ac{PPO} have achieved agile quadrotor flight~\citep{vanspiking}. Nonetheless, these demonstrations were conducted on platforms with comparatively well-understood dynamics and a relatively straightforward mapping between control commands and actuator responses. Moreover, the embedded hardware employed remains substantially more powerful than our target platform (ESP32-S3\footnote{\url{https://www.espressif.com/en/products/socs/esp32}}), which operates under significantly stricter \ac{SWaP} constraints.

\begin{figure}[ht]
    \centering
    \includegraphics[width=0.8\linewidth]{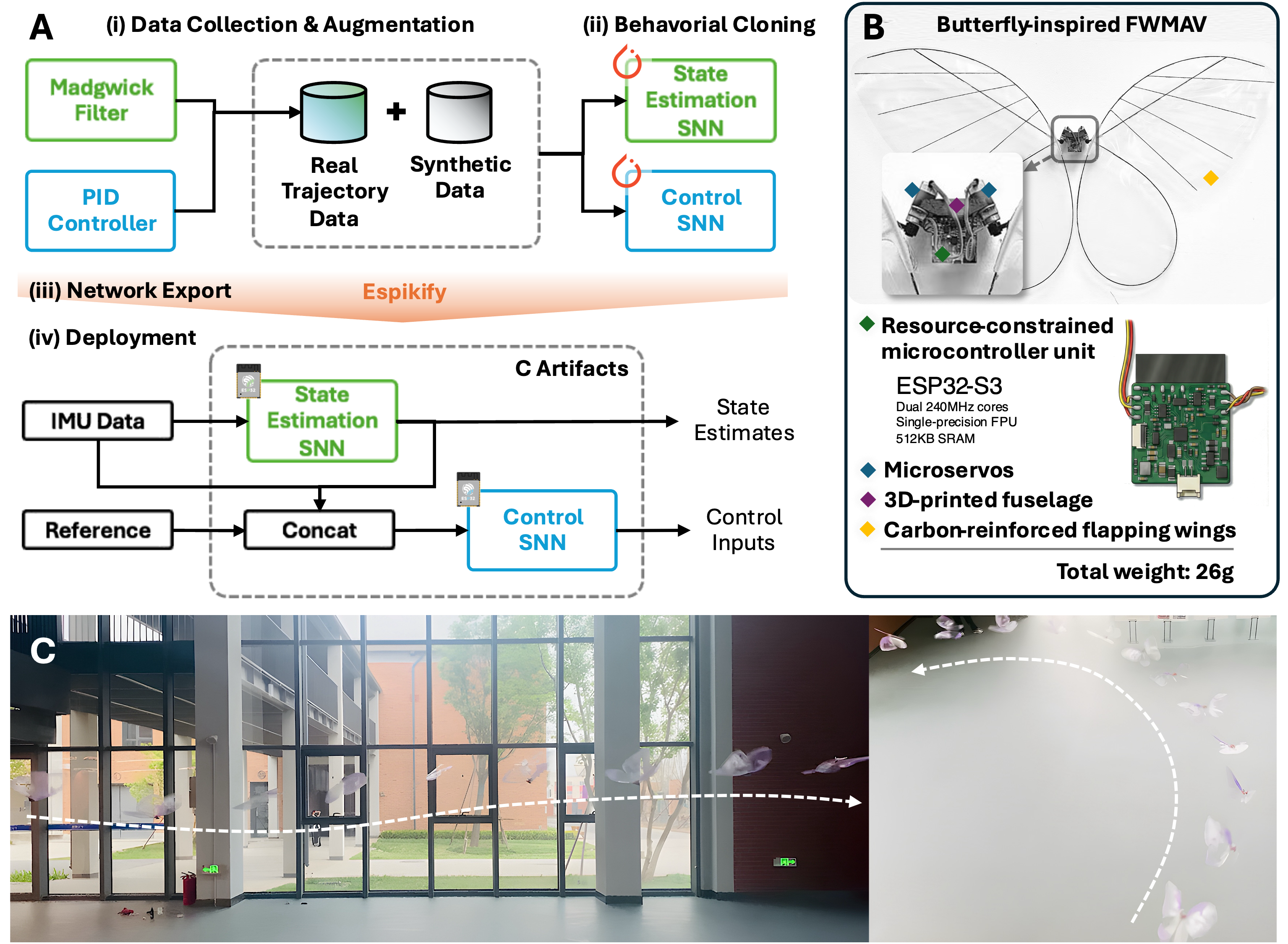}
    \caption{\textbf{Overview of onboard SNN control for a butterfly-inspired FWMAV.} (\textbf{A}) Learning-to-deployment pipeline. Baseline Madgwick estimator and PID controller generate real-flight demonstrations, which are augmented with synthetic data and used for behavioral cloning of state estimation and control SNNs. Custom tool \texttt{Espikify} converts the trained networks into C artifacts for embedded deployment, where IMU measurements and reference commands are processed onboard to produce state estimates and control inputs. (\textbf{B}) The $26\,\mathrm{g}$ FWMAV, comprising microservos, a 3D-printed fuselage, carbon-reinforced flapping wings, and a resource-constrained ESP32-S3 microcontroller. (\textbf{C}) Untethered indoor flight demonstrations of pitch and yaw tracking.}
    \label{fig:teaser}
\end{figure}

To the best of our knowledge, this work presents the first fully onboard \ac{SNN} flight controller for a sub-30~gram butterfly-inspired \ac{FWMAV}, enabling untethered flight via imitation learning  (Fig.~\ref{fig:teaser}A). The robotic platform operates under strict \ac{SWaP} constraints on sensing, computation, and actuation (Fig.~\ref{fig:teaser}B). In addition, its low-aspect-ratio wings amplify unsteady aerodynamic effects compared to many existing flapping-wing designs~\citep{jafferis2019untethered,de2009design,phan2017design,zhang2017design,ramezani2017biomimetic}, while the absence of a tail or auxiliary control surfaces limits actuation authority~\citep{phan2024twist}. With only two wing actuators responsible for generating lift, thrust, and moments, the system is underactuated with respect to its full \ac{6-DoF} in three-dimensional space, rendering control considerably more challenging than for quadrotors or many other flapping-wing platforms. These characteristics, together with the stringent computational and power constraints, motivate the adoption of \acp{SNN}, which provide lightweight, energy-efficient computation capable of sustaining the high control frequencies required for flight on resource-constrained platforms (Fig.~\ref{fig:teaser}C). Our main contributions are as follows:
\begin{itemize}
    \item We introduce a hierarchical neuromorphic control architecture for a sub-30~gram \ac{FWMAV} where \acp{SNN} perform onboard state estimation and attitude control, interfacing with a \ac{CPG} for wing actuation.
    \item We develop and open-source \texttt{Espikify}\footnote{\url{https://github.com/NEAR-the-future/Espikify.git}}, a lightweight tool that converts PyTorch-defined \acp{SNN} into ESP32-ready C code with sub-millisecond inference and minimal memory footprint for low-cost embedded hardware. %The generated runtime minimizes memory footprint and achieves sub-millisecond on-device inference, enabling reproducible deployment on low-cost embedded hardware.
    \item We validate our approach through real-world flight tests and embedded benchmarking, demonstrating reduced inference latency and energy consumption compared to \ac{ANN}-based baselines on identical hardware.
    \item We present the first experimental validation of a fully onboard \ac{SNN} flight control system for a butterfly-inspired \ac{FWMAV} less than 30~gram, achieving stable closed-loop untethered flight on resource-constrained embedded hardware.
\end{itemize}

The remainder of the paper is structured as follows. We first review prior work on learning-based and neuromorphic control for flapping-wing flight in Section~\ref{sec:related_work}. Section~\ref{sec:platform} then introduces the robotic platform and formulates the control problem that motivates the proposed approach. Building on this formulation, Section~\ref{sec:method} first presents the hierarchical \ac{SNN} architecture for state estimation and control, followed by the description of the \texttt{Espikify} conversion tool and the embedded deployment pipeline on the ESP32 that enable fully onboard execution. Experimental validation is provided in Section~\ref{sec:experiments}, including free-flight results and embedded performance benchmarks. Finally, Section~\ref{sec:conclusion} concludes the paper, and Section~\ref{sec:limitations} discusses limitations and directions for future work.

\section{Related Work}
\label{sec:related_work}

\noindent\textbf{Flight control of flapping-wing robots.} Flight control of flapping-wing robots is challenging due to strongly nonlinear, coupled, and time-varying aerodynamics, motivating learning-based methods that reduce reliance on first-principles models. Hsiao \emph{et al.}~\citep{hsiao2025rtmpc_il} achieved agile untethered flight on a 750~mg platform by distilling a robust tube-\ac{MPC} expert into a neural policy, enabling high-speed and high-acceleration maneuvers. Xiong \emph{et al.}~\citep{xiong2023lift} used \ac{RL} to optimize wing kinematics for lift generation on a butterfly-inspired robot, but focused on tethered evaluation and lift maximization rather than closed-loop free-flight control. Cai \emph{et al.}~\citep{cai2025learning} proposed an \ac{RL}-based trajectory tracking framework for bird-inspired flapping-wing robots with validation in simulation only. While these studies highlight the promise of learning for flapping-wing systems, many depend on dense \acp{ANN} and/or limited real-world closed-loop validation, leaving a gap toward high-rate onboard deployment under strict \ac{SWaP} constraints. In contrast, this work addresses fully onboard closed-loop control for untethered flight on resource-constrained hardware.

\noindent\textbf{Neuromorphic control in general robotics.} Neuromorphic computing offers an event-driven alternative to dense \acp{ANN} with potential advantages in latency and energy efficiency for embedded autonomy. Surveys and tutorials summarize \ac{SNN} training methodologies and emphasize the growing interest in robotics applications~\citep{bing2018survey,eshraghian2023training}. Oikonomou \emph{et al.} discuss how \ac{RL} can be adapted to \acp{SNN} and outline potential benefits for control and neuromorphic deployment~\citep{oikonomou2025reinforcement}, however they document that real-time evaluations occur in less than 30\% of the reviewed works, indicating that most studies remain confined to simulation. Closed-loop controllers from the robotics \ac{SNN} literature illustrate these gaps. Bing \emph{et al.}~\citep{bing2019end} trained a multi-layer \ac{RSTDP} controller for snake-like target tracking validated only in the V-REP simulation environment. Jiang \emph{et al.}~\citep{jiang2020target} implemented a \ac{DVS}-driven multi-layer controller for a wheel-less snake robot within the \ac{NRP} simulation framework. Yang \emph{et al.}~\citep{yang2025adaptive} present a real-world validation of an adaptive SNN controller on a simulated vacuum cleaner robot performing wall following, but the controller is designed for a ground vehicle and does not enforce the extreme \ac{SWaP} limits necessary for micro-scale flight. These efforts motivate \acp{SNN} for embedded control, but also imply a lack of evaluations of fully closed-loop, onboard neuromorphic controllers for autonomous platforms under the tight \ac{SWaP} constraints.

\noindent\textbf{\ac{SNN}-based control of \acp{UAV}.} Early work on \ac{SNN}-based flight control was largely restricted to simulation. Clawson \emph{et al.}~\citep{clawson2016spiking} trained a \ac{LIF} \ac{SNN} via reward-modulated Hebbian plasticity to mimic a \ac{LQR} controller for a RoboBee-like platform, and later augmented a linear baseline with an adaptive spiking component to improve robustness under uncertainty~\citep{clawson2017adaptive}—both studies used experimentally informed models but did not demonstrate real-world flight. Recently, aerial demonstrations have focused on rotary-wing platforms. Stagsted \emph{et al.} proposed a spiking implementation of a classical \ac{PID} control structure for quadrotor control, demonstrating a practical pathway for converting conventional controllers into spiking realizations~\citep{stagsted2020towards}. Stroobants \emph{et al.} introduced neuromorphic attitude estimation for quadrotors with embedded deployment considerations~\citep{stroobants2022neuromorphic}, and later demonstrated a modular neuromorphic estimation-and-control stack running at high rates on embedded hardware~\citep{stroobants2025neuromorphic}. Castillo \emph{et al.} advanced this direction by training fully spiking actor--critic policies with PPO for agile quadrotor maneuvers and transferring them to real flight experiments~\citep{vanspiking}. However, existing aerial demonstrations have mostly focused on platforms with well-characterized dynamics and greater actuation authority, a stark contrast to tailless flapping-wing robots, whose strongly coupled, underactuated \ac{6-DoF} dynamics make closed-loop control more demanding. A similar hardware difference exists, as prior work~\citep{stroobants2025neuromorphic} used more capable processors such as Cortex-M7 while we target an ESP32-S3 under stricter \ac{SWaP} constraints\footnote{Different from~\citep{stroobants2025neuromorphic} (Teensy~4.0: 600~MHz ARM Cortex-M7, 1~MB on-chip SRAM), we use ESP32-S3 (240~MHz dual-core, 512~KB on-chip SRAM).}.

% However, existing aerial demonstrations have mostly considered platforms with relatively well-characterized dynamics and greater actuation authority. Tailless flapping-wing robots present a different setting, with strongly coupled and underactuated \ac{6-DoF} dynamics that make closed-loop control more demanding. A similar difference exists on the hardware side. Previous studies have typically used more capable embedded processors such as Cortex-M7~\citep{stroobants2025neuromorphic}, while our implementation targets an ESP32-S3 operating under stricter \ac{SWaP} constraints\footnote{Different from~\citep{stroobants2025neuromorphic} (Teensy~4.0: 600~MHz ARM Cortex-M7, 1~MB on-chip SRAM), we use ESP32-S3 (240~MHz dual-core, 512~KB on-chip SRAM).}.
%\footnote{In~\citep{stroobants2025neuromorphic}, the authors use Teensy~4.0 development board that employs a 600\,MHz ARM Cortex-M7 (NXP i.MX RT1062) with single-precision FPU, DSP extensions, and approximately 1\,MB of on-chip SRAM. In this paper, we target the ESP32-S3-WROOM-1-N16R8, featuring dual 240\,MHz Xtensa LX7 cores with single-precision FPU and 512\,KB of on-chip SRAM, supplemented by external PSRAM. Due to its higher clock rate and larger tightly coupled on-chip memory, the Teensy~4.0 offers significantly greater deterministic floating-point performance for real-time control workloads.}.

\section{Robotic Platform and Problem Formulation}
\label{sec:platform}

A sub-30~gram butterfly-inspired \ac{FWMAV} was fabricated in this work for experimental validation (detailed specifications given in~\citep{gu202626}). As illustrated in Fig.~\ref{fig:teaser}B, the tailless platform features two independently actuated flapping wings, each forewing driven by a micro servo that converts rotary motion into reciprocating flapping at a nominal frequency of $3.25,\mathrm{Hz}$, reproducing the anteromotoric wing actuation observed in butterflies. 
% % Each wing assembly consists of a forewing and hindwing mechanically coupled via the transmission mechanism. 
% The servo directly actuates the forewing, while the hindwing follows through passive mechanical coupling, leading to a synchronized forewing–hindwing motion with a characteristic phase lag. 
Morphologically, the wing structure mimics that of real butterflies: ultra-thin Mylar film forms the membrane, reinforced by carbon-fiber spars that replicate venation patterns, which results in a compliant wing with low aspect ratio. The fuselage integrates a microcontroller unit based on the ESP32-S3 and an \ac{IMU} providing raw inertial measurements for feedback control. %Detailed mechanical and avionics specifications can be found in~\citep{gu202626anonymous}.

Biological flapping flight is often driven by \acp{CPG}, neural circuits that generate rhythmic motor patterns whose characteristics can be modulated by sensory feedback. In our system, rhythmic flapping motion for each wing is produced by a compact sinusoidal oscillator embedded within the closed-loop flight control architecture (Fig.~\ref{fig:methodology}) as
\begin{equation}
\zeta(t) = A \sin\!\bigl(2\pi f t + \varphi\bigr) + o(t),
\label{eq:cpg}
\end{equation}
where $\zeta(t) \in \mathbb{R}$ denotes the commanded wing stroke angle, $A \in \mathbb{R}_{>0}$ the flapping amplitude, $f \in \mathbb{R}_{>0}$ the flapping frequency, and $\varphi \in \mathbb{R}$ the oscillator phase. The term $o(t) \in \mathbb{R}$ is the control input, representing an angle offset generated by the attitude controller that modulates the mean stroke position without altering periodicity. Same (or differential) offsets between the two wings generate pitch (or yaw) moment while preserving smooth flapping kinematics. The oscillator output is mapped to \ac{PWM} commands to actuate the micro servos. 
% Equation~\eqref{eq:cpg} is implemented using a fixed-rate phase accumulator to ensure consistent phase progression. 
During flight, the robot's attitude $(\phi, \theta, \psi)$—roll, pitch, and yaw angle—needs to be estimated from raw \ac{IMU} measurements. Based on these estimates, control inputs should be generated and mapped to the offset input $o(t)$ of the oscillator. The control objective is to track desired pitch and yaw angles in untethered free flight by regulating angle offsets within~\eqref{eq:cpg}. 

\section{Methodology}
% \section{Neuromorphic State Estimation and Control}
\label{sec:method}

\begin{figure}[t] 
\centering 
\includegraphics[width=0.95\linewidth]{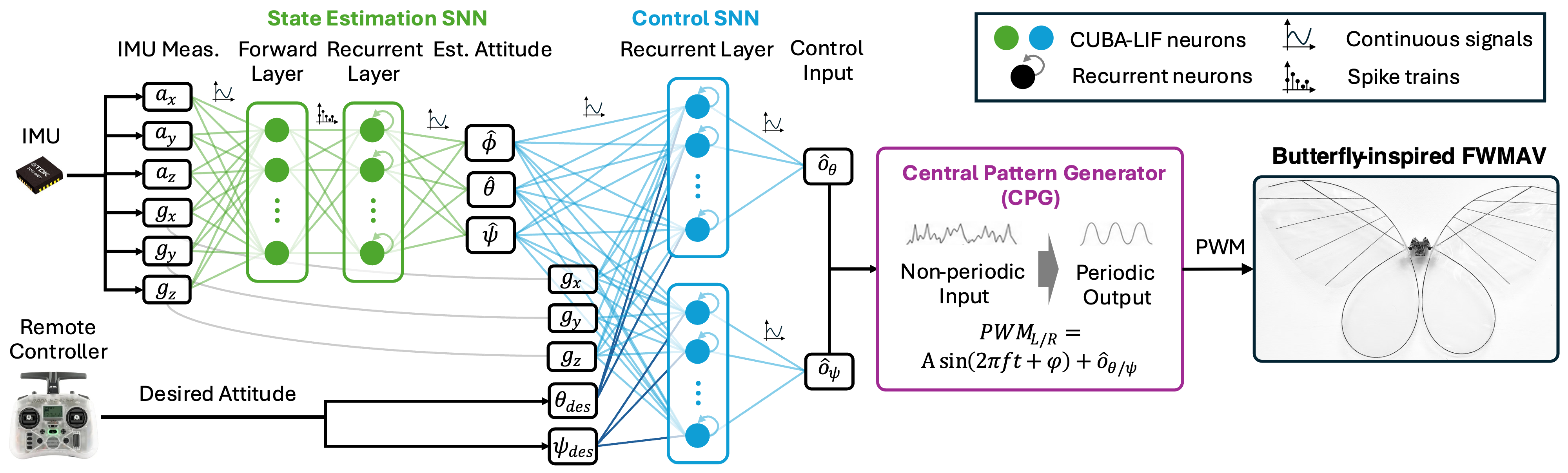} 
\caption{%
\textbf{Hierarchical SNN control architecture for \ac{FWMAV} flight.} The state estimation SNN (green) estimates attitude from raw IMU data. The control SNN (blue) then combines these estimates with gyroscope readings and the desired attitude to generate pitch and yaw offsets, which modulate a CPG (purple) to produce periodic PWM commands for both wing servos.}
\label{fig:methodology} 
\end{figure}

Inspired by neuromorphic quadrotor control~\citep{stroobants2025neuromorphic}, we decompose the onboard control system of our \ac{FWMAV} into spike-based state estimation and control modules (Fig.~\ref{fig:methodology}). The first \ac{SNN} estimates attitude-related states from raw \ac{IMU} data, and the second \ac{SNN} computes control inputs given these estimated states and the references from a remote controller. Notably, our approach differs from~\citep{stroobants2025neuromorphic} in four key aspects. First, wingstroke-induced body undulation (3.25~Hz) introduces periodic components in inertial measurements and attitude signals~\citep{gu202626}, requiring both modules to remain robust to oscillatory inputs. Second, control authority is exerted by modulating a rhythmic actuation pattern via a low-dimensional \ac{CPG} interface rather than direct thrust/torque commands. Third, we target real-time execution on a more resource-constrained microcontroller. Lastly, our state estimator provides full Euler angles required for closed-loop attitude tracking. We next detail the network design, training, and embedded deployment.

\noindent\textbf{Neuron model and training objective.}
Both networks take the form of recurrent \acp{SNN} using the \ac{CUBALIF} neuron model (Fig.~\ref{fig:methodology}). For neuron $i$, the synaptic current $I_i[t]\in\mathbb{R}$ and membrane potential $V_i[t]\in\mathbb{R}$ at time step $t$ evolve as
\begin{align}
I_i[t{+}1] &=
\alpha_i I_i[t]
+ \sum_{j=1}^{N_{\mathrm{ff}}} w_{ij} S_j[t]
+ \sum_{k=1}^{N_{\mathrm{rec}}} \tilde{w}_{ik} S_k[t]
+ U_i[t],
\label{eq:cuba_syn}\\
V_i[t{+}1] &= \beta_i V_i[t] + I_i[t],
\label{eq:cuba_mem}
\end{align}
where $\alpha_i,\beta_i\in(0,1)$ are leak factors, $w_{ij}$ and $\tilde{w}_{ik}$ are feedforward and recurrent weights, $N_{\mathrm{ff}}$ and $N_{\mathrm{rec}}$ are the numbers of feedforward and recurrent presynaptic neurons connected to neuron $i$, $S_j[t],S_k[t]\in\{0,1\}$ are presynaptic spikes, and $U_i[t]$ is an injected input current. Networks are trained independently through \ac{BPTT} using surrogate gradients to address the non-differentiability of spike events~\citep{eshraghian2023training}. Specifically, we use the Arctan surrogate gradient and approximate $\partial S_i[t]/\partial V_i[t]$ during backpropagation with
\begin{equation}
\frac{\partial S_i[t]}{\partial V_i[t]}
\approx
\frac{1}{1+\kappa\bigl(V_i[t]-\Theta_i\bigr)^2},
\label{eq:surrogate}
\end{equation}
where $\Theta_i$ is the firing threshold of neuron $i$, and $\kappa$ is the surrogate-gradient slope coefficient. 
% We set $\kappa=30$ for the state estimation network and $\kappa=60$ for the control network. 
We formulate behavioral cloning~\citep{huangdiffuseloco} as a supervised sequence prediction problem over windows with expert demonstrations generated by the baseline Madgwick estimator and \ac{PID} controller (Appendix~\ref{app:baseline_data}). Given expert targets $\mathbf{y}[t]$ and network estimates $\hat{\mathbf{y}}[t]$, we minimize the Huber loss over windows of length $T$ for its robustness to outliers and differentiability near zero following~\citep{stroobants2025neuromorphic}:
\begin{equation}
\mathcal{L} =
\frac{1}{T}\sum_{t=1}^{T}
\ell_{\mathrm{Huber}}\!\left(\hat{\mathbf{y}}[t],\mathbf{y}[t]\right)
+\lambda\left(1-\rho\!\left(\hat{\mathbf{y}}[1{:}T],\mathbf{y}[1{:}T]\right)\right),
\label{eq:loss}
\end{equation}
where $\rho(\cdot,\cdot)$ is the Pearson correlation computed over the sequence and $\lambda$ weights the correlation term. We use the elementwise Huber loss with $\delta=1.0$ and exclude the first 100\, time steps of each sequence from the loss to reduce sensitivity to hidden-state initialization. The full loss definition and training details are provided in Appendix~\ref{app:snn_training_and_io}.

\noindent\textbf{State estimation network.} The state estimation \ac{SNN} maps raw six-axis IMU measurements to a three-dimensional state estimate (Fig.~\ref{fig:methodology}). We exclude magnetometer measurements due to indoor magnetic disturbances that could induce large heading biases. Consequently, the network is supervised using only roll $\phi$, pitch $\theta$, and yaw rate $\dot{\psi}$, rather than absolute yaw, because yaw angle is not globally observable without a magnetometer and drifts over time due to gyroscope bias accumulation. Training targets are generated by an onboard six-axis Madgwick filter, which fuses linear accelerations from accelerometer and angular rates from gyroscope. The network architecture consists of a 150-neuron feedforward spiking layer, followed by a 150-neuron recurrent spiking layer, and a linear readout that produces the three outputs.

\noindent\textbf{Control network.} We consider two variants of \ac{SNN} for control. Both receive the same input of references, estimated states, and raw gyroscope measurements. Each controller is trained via behavioral cloning to reproduce the outputs of the baseline \ac{PID} controller. The network architecture consists of a recurrent spiking layer with 130 neurons, followed by a linear readout.
\begin{itemize}
    \item[\textbf{(1)}] \textbf{CPG-agnostic control \ac{SNN}.}
This variant directly predicts left and right servo commands:
\begin{equation}
\hat{\mathbf{u}}_{\mathrm{PWM}}[t] =
\begin{bmatrix}
\widehat{\mathrm{PWM}}_{L}[t]\\
\widehat{\mathrm{PWM}}_{R}[t]
\end{bmatrix}
\in \mathbb{R}^{2\times1},
\label{eq:pwm_control}
\end{equation}
where the entries denote the predicted \ac{PWM} commands\footnote{Here, PWM command denotes the pulse duration (in $\mu s$) of the $50\,\mathrm{Hz}$ servo signal, not the full waveform.} sent to the left and right servo driver. This formulation is the least constrained controller, as it must learn both the rhythmic flapping pattern and the feedback correction directly from data. 

\item[\textbf{(2)}] \textbf{CPG-aware control \ac{SNN}.} This variant predicts angle offsets that modulate the \ac{CPG}~\eqref{eq:cpg} (Fig.~\ref{fig:methodology}). Based on control effectiveness~\citep{gu202626}, we consider two instances of how the learned offset is applied to the left and right wings. The \textit{pitch-offset} instance outputs a scalar $\hat{o}_{\theta}[t]\in\mathbb{R}$ applied identically to both wings:
\begin{equation}
o_{L}(t) = o_{R}(t) = \hat{o}_{\theta}[t],
\label{eq:pitch_offset}
\end{equation}
where $o_{L}(t), o_{R}(t) \in \mathbb{R}$ are the offset inputs to the \ac{CPG}~\eqref{eq:cpg}. This produces longitudinal control by shifting the mean stroke symmetrically while preserving rhythmic flapping. The \textit{yaw-offset} instance outputs a scalar $\hat{o}_{\psi}[t]\in\mathbb{R}$ applied with opposite signs:
\begin{equation}
o_{L}(t) = \hat{o}_{\psi}[t], \quad o_{R}(t) = -\hat{o}_{\psi}[t],
\label{eq:yaw_offset}
\end{equation}
generating a yawing moment while maintaining smooth rhythmic wing kinematics.
\end{itemize}

\noindent\textbf{Embedded deployment.}
After training, both the state estimation and control \acp{SNN} are converted from PyTorch into static C artifacts for execution on the ESP32 using \texttt{Espikify} (Fig.~\ref{fig:teaser}A). This tool exports learned weights, neuron parameters, and the static network configuration as compile-time headers. The deployed model preserves the training-time cascaded structure. At each timestep $t$, the state estimator executes first, producing an intermediate attitude estimate $\hat{\mathbf{s}}[t]\in\mathbb{R}^{d_s}$. This estimate is concatenated with reference and measurement signals $\mathbf{r}[t]\in\mathbb{R}^{d_r}$ to form the full controller input $\mathbf{x}_{\mathrm{ctrl}}[t] = [\mathbf{r}[t]^\top, \hat{\mathbf{s}}[t]^\top ]^\top \in \mathbb{R}^{d_r + d_s}$,
% \begin{equation}
% \mathbf{x}_{\mathrm{ctrl}}[t] = \begin{bmatrix} \mathbf{r}[t] \\ \hat{\mathbf{s}}[t] \end{bmatrix} \in \mathbb{R}^{d_r + d_s},
% \label{eq:embedded_cascade_input}
% \end{equation}
where $d_r$ and $d_s$ are the respective dimensionalities. The controller then maps $\mathbf{x}_{\mathrm{ctrl}}[t]$ to the final control command. In the embedded flight stack, the \ac{SNN}-based estimator and controller directly replace the original Madgwick filter and \ac{PID} controller, while all other sensing, timing, and actuation interfaces remain unchanged.

\section{Experimental Results}
\label{sec:experiments}
\vspace{-0.6em}

\begingroup
\setlength{\intextsep}{0pt}
\setlength{\columnsep}{0.8em}

\noindent
\begin{wrapfigure}[20]{r}{0.45\textwidth}
\vspace{-0.3em}
\centering
\includegraphics[width=\linewidth]{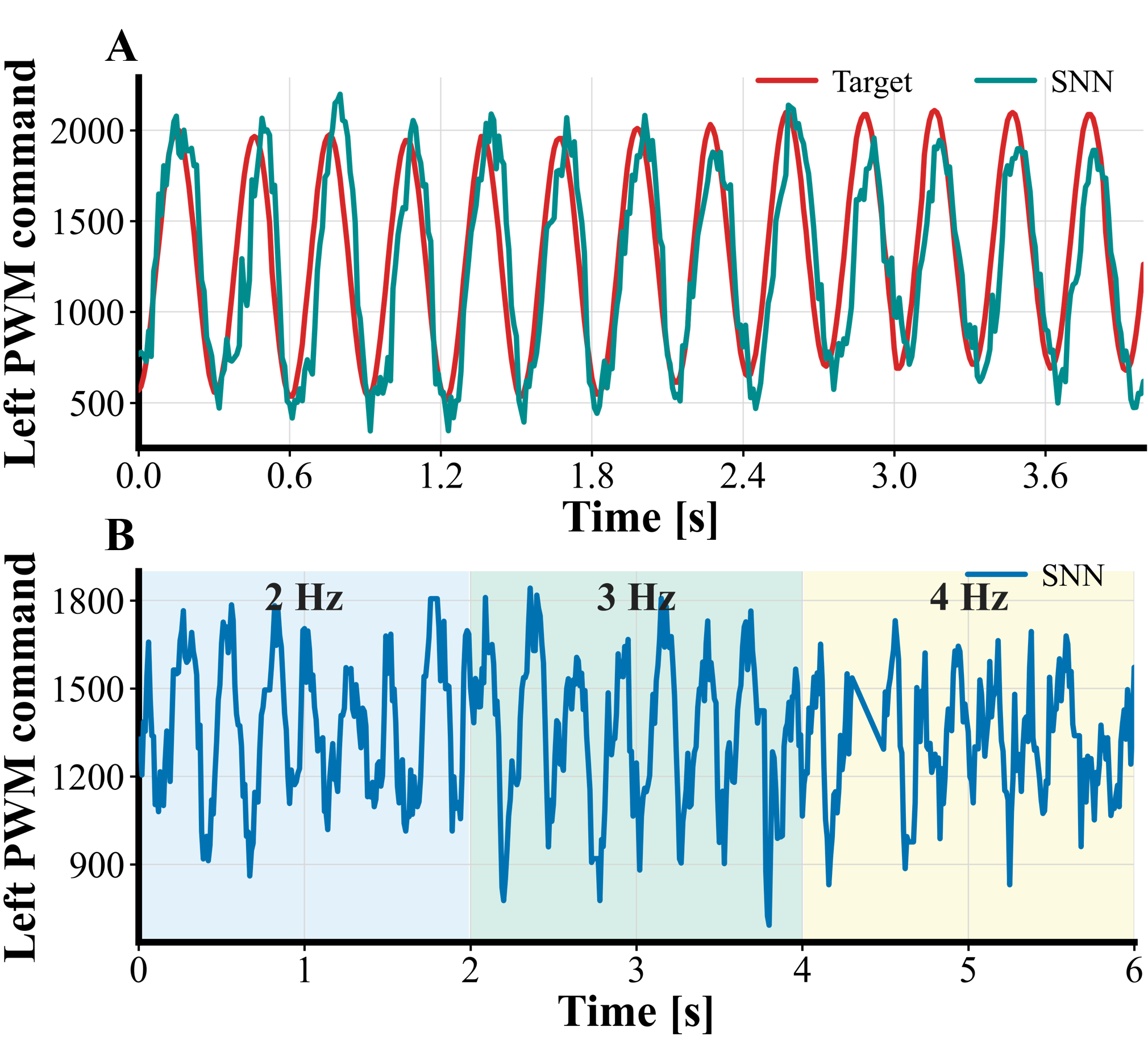}
\caption{%
\textbf{Generalization capability of CPG-agnostic controller.}
(\textbf{A}) Offline replay on a held-out test set near the training regime (3.25~Hz).
(\textbf{B}) At 2/3/4~Hz frequency sweeps, irregular PWM predictions occur outside the training regime.
}
\label{fig:pwm_exp_b}
\vspace{-1em}
\end{wrapfigure}
\textbf{Dataset preparation.}
Training data were collected during indoor free flight using the baseline flight stack as demonstration. Each flight log includes six-axis \ac{IMU} data, Madgwick estimates, references, and baseline \ac{PID} offset commands. Real-flight data, totaling approximately 10~min, were segmented into overlapping windows of length $T=2500$ with stride 400. Since imitation learning only covers the state space visited during stable operation~\citep{ross2011reduction}, the learned subnetworks may generalize poorly to unseen conditions. To broaden the training distribution, we generated approximately 150~min of synthetic flight data (Fig.~\ref{fig:teaser}A). See Appendix~\ref{app:baseline_data} for additional details.

\noindent\textbf{CPG-agnostic versus CPG-aware control.} Offline replay (Fig.~\ref{fig:pwm_exp_b}A) shows that the CPG-agnostic controller reproduces expert PWM commands near the nominal 3.25~Hz training condition. However, in flight tests, direct PWM imitation becomes irregular at other frequencies, indicating limited generalization to unseen body undulation (Fig.~\ref{fig:pwm_exp_b}B). Together with the subsequent results, this confirms the \ac{CPG}-aware controller as the better choice. Delegating the rhythmic wing stroke to a deterministic \ac{CPG} reduces the search space and simplifies imitation learning, as the \ac{SNN} only needs to learn corrective offsets. This separation preserves physically consistent wing motion and improves robustness beyond the training regime.

\endgroup

\begin{figure}[ht]
\vspace{-1em}
    \centering
    \includegraphics[width=\linewidth]{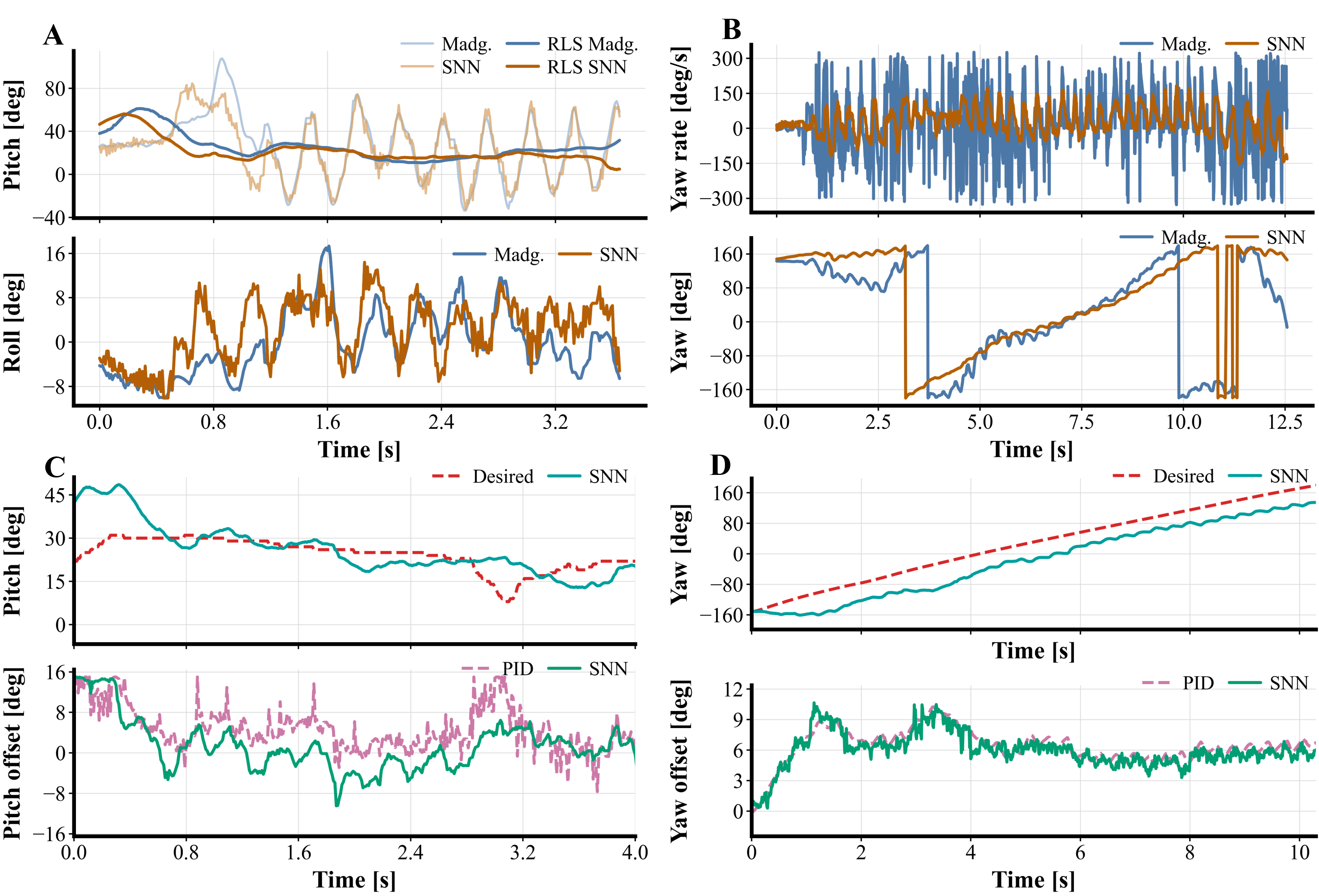}
    \caption{%
    \textbf{Attitude estimation and control performance of the deployed \acp{SNN} from free flight.} (\textbf{A, B}) Comparison of SNN and Madgwick filter attitude estimates for pitch (A) and yaw (B). (\textbf{C, D}) Filtered pitch (C) and yaw (D) tracking. Upper subplots: SNN versus Madgwick attitude estimates. Lower subplots: learned SNN offset commands (solid) versus reconstructed PID baselines (dashed).}
    \label{fig:state_control}
\end{figure}

\begingroup
\setlength{\intextsep}{0pt}
\setlength{\columnsep}{0.8em}

\begin{wraptable}[14]{r}{0.58\textwidth}
\vspace{0.1em}
\centering
\scriptsize
\caption{\textbf{Hierarchical SNN performance in free flight.} Values are mean and standard deviation of segment-level RMSE across five independent flight trialsIt. Pitch estimates are RLS-filtered.}
\label{tab:controller_results}
\begin{tabular}{@{}lllcc@{}}
\toprule
\textbf{Controller} & \textbf{Task} & \textbf{Signal} & \textbf{RMSE (°)} & \textbf{Std. (°)} \\
\midrule
\multirow{4}{*}{Pitch-offset}
& \multirow{2}{*}{Estimation}
& Pitch & 6.96 & 1.58 \\
& & Roll  & 5.22 & 1.18 \\
\cmidrule(lr){2-5}
& \multirow{2}{*}{Control}
& Pitch tracking & 6.32 & 1.52 \\
& & Pitch offset & 4.55 & 0.38 \\
\midrule
\multirow{5}{*}{Yaw-offset}
& \multirow{3}{*}{Estimation}
& Pitch & 5.61 & 3.04 \\
& & Roll & 4.53 & 1.77 \\
& & Yaw & 6.43 & 3.72 \\
\cmidrule(lr){2-5}
& \multirow{2}{*}{Control}
& Yaw tracking & 8.88 & 1.17 \\
& & Yaw offset & 0.86 & 0.16 \\
\bottomrule
\end{tabular}
\vspace{-1.0em}
\end{wraptable}

\noindent\textbf{Real-world validation.}
Table~\ref{tab:controller_results} reports estimation and control accuracy using segment-level \ac{RMSE} from free-flight data. Estimation errors are computed with respect to Madgwick filter estimates, while control errors are evaluated against commanded attitudes or synchronized \ac{PID} offset references reconstructed from the same logs used to collect \ac{SNN} outputs. The resulting \ac{SNN} closely matches the baseline estimation and control behavior for both pitch- and yaw-offset instances. Figure~\ref{fig:state_control} shows one of the representative flight traces. The deployed \ac{SNN} captures the dominant pitch, roll, and yaw dynamics required for feedback control, including yaw reconstructed from the estimated yaw rate. Due to body undulation from flapping motion, raw pitch estimates contain periodic oscillations. We therefore report both raw and \ac{RLS}-filtered pitch, with the latter used by our \ac{PID} controller for smoother feedback. The learned pitch and yaw offset commands follow the corrective trends of the reconstructed \ac{PID} baselines, supporting stable tracking in untethered flight.

\endgroup

\begin{wrapfigure}[15]{r}{0.45\textwidth}
\vspace{-0.5em}
\centering
\includegraphics[width=\linewidth]{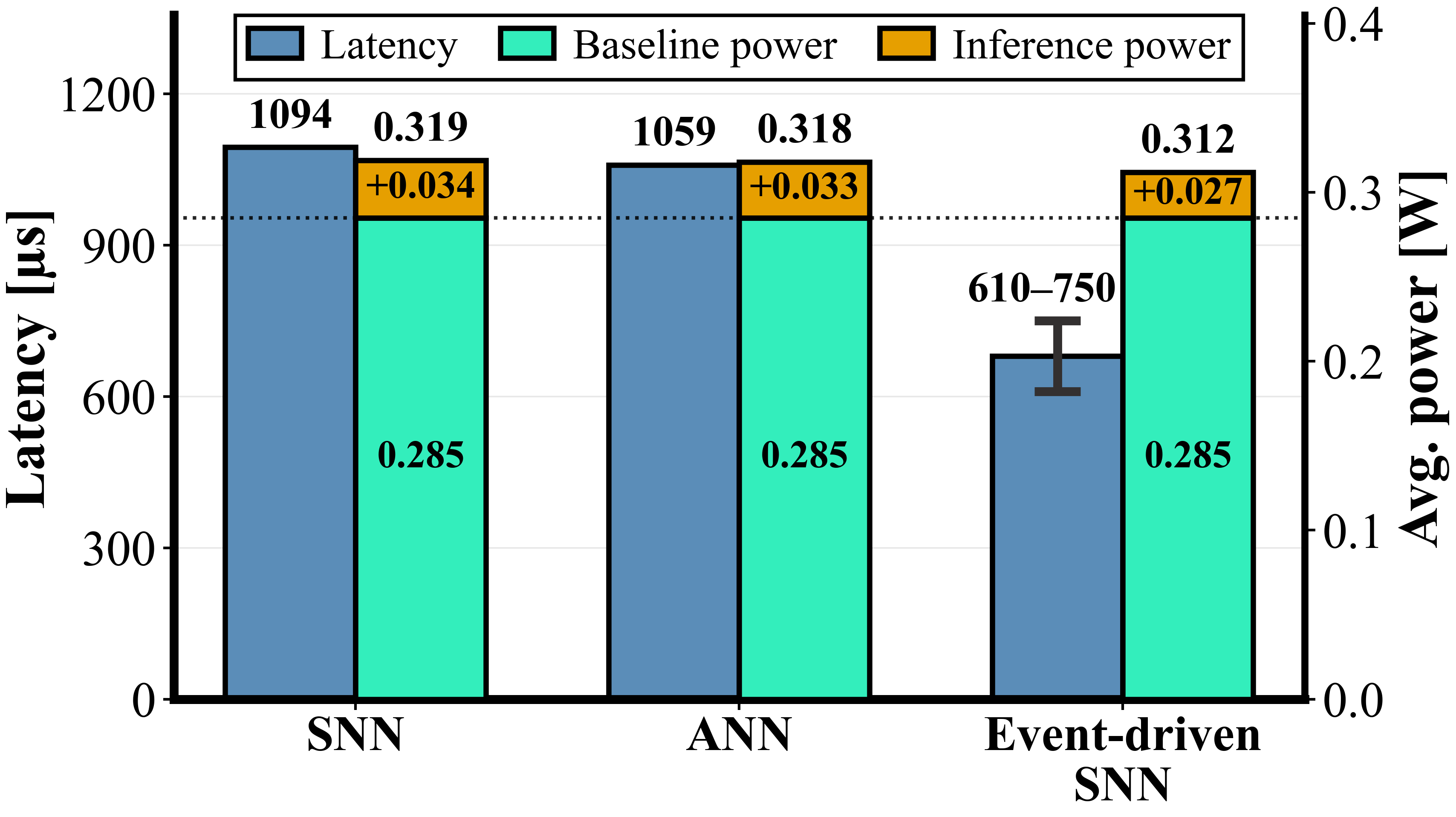}
\caption{%
\textbf{Latency and power of SNN, ANN, and event-driven SNN for single onboard inference on ESP32-S3.} Power bars are total board power with inference-induced increments shown above the baseline.}
\label{fig:esp32_latency_power}
\vspace{-1.0em}
\end{wrapfigure}

\noindent\textbf{Computational efficiency.}
We benchmarked the hierarchical onboard control inference on ESP32, with each $100\,\mathrm{Hz}$ \ac{IMU} interrupt triggering one inference step (Appendix~\ref{app:embedded_benchmark}). Figure~\ref{fig:esp32_latency_power} shows our event-driven \ac{SNN} implementation (Appendix~\ref{app:event_driven_snn}) reduces latency by $36\%$ and inference-induced power by $18\%$ relative to the unoptimized dense-float \ac{ANN} baseline (Appendix~\ref{app:ann}). Unlike neuromorphic chips that natively support event-driven processing, ESP32 lacks dedicated hardware. Consequently, a conventionally implemented \ac{SNN} (without event-driven optimization) can perform only comparably to the \ac{ANN}. We also report a supplementary ablation study comparing scalar and \ac{SIMD} implementations in Appendix~\ref{app:embedded_benchmark}. The results show that the \ac{ANN} and dense \ac{SNN} benefit more from \ac{SIMD} than the event-driven \ac{SNN} does, while their inference latency remains higher than that of the event-driven \ac{SNN}.

\section{Conclusion}
\label{sec:conclusion}

This paper demonstrates energy-efficient onboard \ac{SNN} control for a two-winged, tailless \ac{FWMAV} operating under strict \ac{SWaP} constraints. The \ac{CPG}-aware \ac{SNN} controller preserves rhythmic flapping while learning only pitch and yaw angle offsets, achieving real-time flight control on an ESP32. Furthermore, our custom, open-sourced tool \texttt{Espikify} lowers the barrier to deploying \acp{SNN} on low-cost, widely available microcontrollers, broadening access to neuromorphic control in robotics.

%===============================================================================

\section{Limitations}
\label{sec:limitations}

Several limitations point to our future work. First, yaw angle is reconstructed through open-loop integration of predicted yaw rate, which allows bias to accumulate; visual-inertial feedback could compensate for this drift. Second, our SNNs now have a slightly larger memory footprint than the ANN baselines (Table~\ref{tab:deployed_footprint}) because they store neuron states in addition to synaptic weights, though this overhead can be reduced with lighter architectures or neuromorphic hardware. We have not yet systematically compared against vector-optimized ANN or quantized baselines, both of which are worthwhile controlled comparisons for future study. Lastly, behavioral cloning inherits teacher bias, and our out-of-distribution evaluation currently only varies flapping frequency, leaving wind gusts and outdoor conditions for subsequent work with online adaptation and broader testing.

\clearpage
% The acknowledgments are automatically included only in the final and preprint versions of the paper.
\acknowledgments{
% If a paper is accepted, the final camera-ready version will (and probably should) include acknowledgments. All acknowledgments go at the end of the paper, including thanks to reviewers who gave useful comments, to colleagues who contributed to the ideas, and to funding agencies and corporate sponsors that provided financial support.
% \ifpreprint{}
Weibin Gu acknowledges support from the China Postdoctoral Science Foundation under Grant Number 2025M781650. This work was also partly supported by Xinchen Qihang Inc., PR China.}

%===============================================================================

\bibliography{reference}  % .bib

\appendix
\clearpage
\section*{Appendices}

\addcontentsline{toc}{section}{Appendices}
\section{Expert Demonstration and Dataset Preparation}
\label{app:baseline_data}

Real-flight demonstrations were collected with the baseline embedded flight stack and used as expert demonstrations for \ac{SNN} training. The real-world training data comprise approximately 10 min of \ac{PID}-controlled untethered free flight, including 33 yaw and 45 pitch flight segments with durations ranging from 2 to 14 s. Forward speed and altitude were not independently prescribed because the onboard controller regulates only pitch and yaw, yet the flights cover natural variations within the stable operating envelope. %Table~\ref{tab:controller_results} and Figure~\ref{fig:state_control} average five independent flight trials, demonstrating repeatable estimation and control.

Pitch and yaw tracking are defined with respect to commanded references. Let $\theta_{\mathrm{ref}}(t)\in\mathbb{R}$ and $\psi_{\mathrm{ref}}(t)\in\mathbb{R}$ denote the pitch and yaw references at time $t$, and let $\theta(t)\in\mathbb{R}$ and $\psi(t)\in\mathbb{R}$ denote the corresponding feedback signals. The tracking errors are
\begin{equation}
e_\theta(t) = \theta_{\mathrm{ref}}(t) - \theta(t),
\qquad
e_\psi(t) =
\operatorname{wrap}_{[-180^\circ,180^\circ)}
\!\left(\psi_{\mathrm{ref}}(t) - \psi(t)\right).
\label{eq:pid_errors}
\end{equation}
Here, $e_\theta(t)\in\mathbb{R}$ and $e_\psi(t)\in\mathbb{R}$ are the pitch and yaw tracking errors, respectively, and $\operatorname{wrap}_{[-180^\circ,180^\circ)}(\cdot)$ maps the yaw error to the interval $[-180^\circ,180^\circ)$. The expert control command for channel $i\in\{\theta,\psi\}$ is computed as
\begin{equation}
u_i(t) = K_{P,i} e_i(t)
+ K_{I,i} \int_0^t e_i(\tau)\, d\tau
+ K_{D,i} \frac{d e_i(t)}{dt},
\qquad i \in \{\theta,\psi\},
\label{eq:pid_law}
\end{equation}
where $u_i(t)\in\mathbb{R}$ is the expert control command, $e_i(t)\in\mathbb{R}$ is the corresponding tracking error, $\tau$ is the integration variable, and $K_{P,i}$, $K_{I,i}$, and $K_{D,i}$ are scalar proportional, integral, and derivative gains. The pitch-offset controller uses $K_{P,\theta}=0.6$, $K_{I,\theta}=0.55$, and $K_{D,\theta}=0.05$, whereas the yaw-offset controller uses $K_{P,\psi}=0.15$, $K_{I,\psi}=0.0$, and $K_{D,\psi}=0.0$.

Before generating synthetic data, the offline data generation procedure was validated on real flight logs by replaying the recorded gyroscope and accelerometer measurements through the same baseline estimation, filtering, and control pipeline used onboard. The replayed attitude and control outputs were compared with the labels stored during flight, confirming that the offline procedure reproduces the real flight processing pipeline. After this validation, synthetic trajectories were generated using low-frequency attitude references, randomized offsets, slow sweeps, tracking delays, transient recovery, and saturation cases. Synthetic \ac{IMU} measurements were produced from these trajectories, and the validated offline baseline pipeline was used to generate the corresponding expert labels.

The yaw-rate target for state estimation is obtained from the baseline yaw estimate by first-order backward finite difference
\begin{equation}
\dot{\psi}[t]
\approx
\frac{\psi[t]-\psi[t-1]}{\Delta t_{\mathrm{ms}}},
\qquad
\text{where }\Delta t_{\mathrm{ms}}=10\,\mathrm{ms}.
\label{eq:yaw_rate_app}
\end{equation}
Here, $\dot{\psi}[t]\in\mathbb{R}$ is the yaw-rate target at timestep $t$, $\psi[t]\in\mathbb{R}$ is the baseline yaw estimate, and $\Delta t_{\mathrm{ms}}\in\mathbb{R}$ is the sampling interval expressed in milliseconds. Yaw rate is used as the supervised target because absolute yaw is not globally observable from six-axis inertial sensing alone.

All models are trained in a scaled signal space. For raw input $\mathbf{x}[t]\in\mathbb{R}^{D}$ and target $\mathbf{y}[t]\in\mathbb{R}^{O}$, the scaled quantities are
\begin{equation}
\tilde{\mathbf{x}}[t]
=
\mathbf{c}_x \odot \mathbf{x}[t],
\qquad
\tilde{\mathbf{y}}[t]
=
\mathbf{c}_y \odot \mathbf{y}[t],
\label{eq:scaling_app}
\end{equation}
where $\tilde{\mathbf{x}}[t]\in\mathbb{R}^{D}$ and $\tilde{\mathbf{y}}[t]\in\mathbb{R}^{O}$ are the scaled input and target vectors, $\mathbf{c}_x\in\mathbb{R}^{D}$ and $\mathbf{c}_y\in\mathbb{R}^{O}$ are fixed channel-wise scaling vectors, and $\odot$ denotes elementwise multiplication. The same scaling is used during training, export, and embedded inference, and predicted outputs are converted back to physical units by elementwise division by the corresponding target scale.

\section{SNN Architecture and Training Details}
\label{app:snn_training_and_io}

\paragraph{Spiking dynamics.}
For neuron $i$ at discrete timestep $t$, a binary spike $S_i[t]\in\{0,1\}$ is emitted when the membrane potential $V_i[t]\in\mathbb{R}$ crosses the learned threshold $\Theta_i\in\mathbb{R}$:
\begin{equation}
S_i[t] =
H\!\left(V_i[t]-\Theta_i\right),
\label{eq:spike_app}
\end{equation}
where $H(\cdot)$ is the Heaviside step function. After a spike, the recurrent voltage update applies a hard reset by suppressing the contribution of the previous membrane potential:
\begin{equation}
V_i[t{+}1]
=
\beta_i V_i[t]\bigl(1-S_i[t]\bigr)
+
I_i[t].
\label{eq:hard_reset_app}
\end{equation}
Here, $V_i[t{+}1]\in\mathbb{R}$ is the updated membrane potential, $\beta_i\in(0,1)$ is the membrane leak factor, and $I_i[t]\in\mathbb{R}$ is the synaptic current of neuron $i$.

Continuous sensor and reference signals are injected directly into the spiking layers rather than being converted into spike trains. For an input vector $\mathbf{x}[t]\in\mathbb{R}^{d}$, the injected current vector is
\begin{equation}
\mathbf{j}[t]
=
\mathbf{W}_{\mathrm{in}}\mathbf{x}[t],
\label{eq:input_injection_app}
\end{equation}
where $\mathbf{j}[t]\in\mathbb{R}^{N}$ is the layer input current, $N$ is the number of neurons in the layer, and $\mathbf{W}_{\mathrm{in}}\in\mathbb{R}^{N\times d}$ is the input projection matrix. In recurrent layers, this input term is combined with the recurrent contribution from the previous spike vector. All input projections are implemented without additive biases.

Each subnetwork produces continuous-valued outputs from the spike vector of its final spiking layer. Let $\mathbf{z}[t]\in\{0,1\}^{M}$ denote this spike vector, where $M$ is the number of neurons in the final spiking layer. The decoded output is
\begin{equation}
\hat{\mathbf{y}}[t]
=
\mathbf{W}_{\mathrm{out}}\mathbf{z}[t],
\label{eq:readout_app}
\end{equation}
where $\hat{\mathbf{y}}[t]\in\mathbb{R}^{O}$ is the output vector, $O$ is the number of output channels, and $\mathbf{W}_{\mathrm{out}}\in\mathbb{R}^{O\times M}$ is the readout matrix. The state estimation network uses the spike vector of its recurrent hidden layer, while each control network uses the spike vector of its single recurrent layer. The readout is also bias-free to reduce memory use and per-step arithmetic on the ESP32.

\paragraph{Loss function.}
The state estimation and control networks are trained with separate supervised objectives. For a scalar prediction error $e\in\mathbb{R}$, we use the Huber loss with threshold $\delta=1.0$:
\begin{equation}
\ell_{\mathrm{Huber}}(e)
=
\begin{cases}
\frac{1}{2}e^2, & |e|\le \delta,\\
\delta |e|-\frac{1}{2}\delta^2, & \mathrm{otherwise}.
\end{cases}
\label{eq:huber_app}
\end{equation}
Here, $\ell_{\mathrm{Huber}}(e)\in\mathbb{R}$ is the scalar loss value. Let $\hat{\mathbf{Y}},\mathbf{Y}\in\mathbb{R}^{T\times B\times O}$ denote the predicted and target output sequences, where $T$ is the sequence length, $B$ is the batch size, and $O$ is the number of output channels. We use $\mathcal{H}_{\delta}(\hat{\mathbf{Y}},\mathbf{Y};t_0{:}T)$ to denote the mean elementwise Huber loss over timesteps $t=t_0,\dots,T-1$, all batch elements, and all output channels. The state estimation network loss excludes the first $100$ timesteps to reduce sensitivity to recurrent-state initialization:
\begin{equation}
\mathcal{L}_{\mathrm{est}}
=
\mathcal{H}_{\delta}
\left(
\hat{\mathbf{Y}},\mathbf{Y};
100{:}T
\right).
\label{eq:loss_est_app}
\end{equation}
The control networks are trained over the full sequence using a Huber term and a Pearson-correlation term:
\begin{equation}
\mathcal{L}_{\mathrm{ctrl}}
=
\mathcal{H}_{\delta}
\left(
\hat{\mathbf{Y}},\mathbf{Y};
0{:}T
\right)
+
\lambda
\left(
1-\rho(\hat{\mathbf{Y}},\mathbf{Y})
\right),
\qquad
\lambda=0.5,
\label{eq:loss_ctrl_app}
\end{equation}
where $\mathcal{L}_{\mathrm{est}},\mathcal{L}_{\mathrm{ctrl}}\in\mathbb{R}$ are scalar training losses, $\lambda\in\mathbb{R}$ is the correlation-loss weight, and $\rho(\hat{\mathbf{Y}},\mathbf{Y})\in\mathbb{R}$ is the Pearson correlation computed over the temporal dimension for each output channel and averaged over the batch and output dimensions.

\paragraph{Embedded execution.}
The exported models are executed inside the embedded flight stack at $100\,\mathrm{Hz}$. The implementation uses fixed-size buffers allocated during initialization, so inference does not require dynamic memory allocation during flight. The firmware uses FreeRTOS to separate time-critical control from lower-priority communication and logging. A high-priority periodic task executes sensing, preprocessing, estimation, and control, while servo updates are handled separately to maintain stable \ac{PWM} timing. Telemetry and logging run at lower priority to reduce interference with the control loop. At startup, gyroscope bias is estimated from stationary measurements and subtracted from subsequent angular-rate samples.

For yaw offset control, the controller requires a yaw angle input during deployment. This signal is reconstructed from the \ac{SNN} predicted yaw rate. After optional initial offset correction, the reconstructed yaw angle is updated using trapezoidal integration:
\begin{equation}
\tilde{\psi}[t]
\approx
\tilde{\psi}[t-1]
+
\frac{\Delta t}{2}
\left(
\widehat{\dot{\psi}}[t-1]
+
\widehat{\dot{\psi}}[t]
\right),
\label{eq:yaw_integration_app}
\end{equation}
where $\tilde{\psi}[t]$ is the reconstructed yaw angle, $\widehat{\dot{\psi}}[t]$ is the predicted yaw rate, and $\Delta t$ is the control-loop period. Before deployment, the exported C artifacts are validated offline by running it on recorded time-series inputs and comparing its outputs with the corresponding PyTorch model outputs.

\section{Event-driven SNN Implementation}
\label{app:event_driven_snn}

The event-driven \ac{SNN} implementation modifies only the spike-mediated synaptic propagation. Dense real-valued input projections are unchanged, while hidden and readout projections driven by binary spikes are evaluated only for active presynaptic neurons. Let a synaptic connection matrix denoted by $\mathbf{W}\in\mathbb{R}^{m\times n}$, where $m$ is the number of postsynaptic neurons and $n$ is the number of presynaptic neurons. The presynaptic spike vector at timestep $t$ is denoted by $\mathbf{s}[t]\in\{0,1\}^{n}$, where $s_j[t]\in\{0,1\}$ is the spike state of presynaptic neuron $j$. The active presynaptic spike set is
\begin{equation}
\mathcal{A}[t]=\{j \in \{1,\dots,n\} \mid s_j[t]=1\},
\label{eq:active_spike_set}
\end{equation}
where $\mathcal{A}[t]$ contains the indices of active presynaptic neurons at timestep $t$. Instead of computing the dense product $\mathbf{W}\mathbf{s}[t]$, the event-driven update accumulates only the columns of $\mathbf{W}$ associated with active spikes 
\begin{equation}
p_{\mathrm{syn},i}[t]
=
\sum_{j\in\mathcal{A}[t]} W_{ij},
\qquad i=1,\dots,m .
\label{eq:event_driven_synaptic_update}
\end{equation}
The scalar $p_{\mathrm{syn},i}[t]\in\mathbb{R}$ is the accumulated synaptic input to postsynaptic neuron $i$, and $W_{ij}\in\mathbb{R}$ is the synaptic weight from presynaptic neuron $j$ to postsynaptic neuron $i$. This reduces the spike-mediated propagation cost when the number of active spikes is small. On the ESP32 hardware, the benefit is somewhat constrained by sparse-index overhead, reduced memory locality, and the remaining dense input projections. However, as shown in Fig.~\ref{fig:esp32_latency_power}, such implementation is effective in reducing inference latency relative to the dense \ac{SNN} while preserving the same deployed structure.

\section{ANN Implementation}
\label{app:ann}

For comparison with the proposed neuromorphic state estimation and control pipeline, we implement an unoptimized dense-float \ac{ANN} baseline using bias-free recurrent neural networks with \ac{ReLU} activations. The \ac{ANN} and \ac{SNN} models use the same input and output signals, supervised objectives, optimizer settings, and training schedule. The state estimation network consists of a $150$-unit feedforward layer, a $150$-unit recurrent hidden layer, and a linear readout. The control network consists of a single $130$-unit recurrent hidden layer followed by a linear readout. For both networks, the recurrent update and output decoding are described as
\begin{equation}
\mathbf{h}[t]
=
\operatorname{ReLU}\!\left(
\mathbf{W}_{\mathrm{in}}\mathbf{p}[t]
+
\mathbf{W}_{\mathrm{rec}}\mathbf{h}[t-1]
\right),
\qquad
\hat{\mathbf{y}}[t]
=
\mathbf{W}_{\mathrm{out}}\mathbf{h}[t],
\label{eq:ann_recurrent_update}
\end{equation}
where $\mathbf{p}[t]\in\mathbb{R}^{d_{\mathrm{in}}}$ is the recurrent-layer input, $\mathbf{h}[t]\in\mathbb{R}^{n_{\mathrm{hid}}}$ is the hidden-state vector, and $\hat{\mathbf{y}}[t]\in\mathbb{R}^{d_{\mathrm{out}}}$ is the output vector. The matrices $\mathbf{W}_{\mathrm{in}}\in\mathbb{R}^{n_{\mathrm{hid}}\times d_{\mathrm{in}}}$, $\mathbf{W}_{\mathrm{rec}}\in\mathbb{R}^{n_{\mathrm{hid}}\times n_{\mathrm{hid}}}$, and $\mathbf{W}_{\mathrm{out}}\in\mathbb{R}^{d_{\mathrm{out}}\times n_{\mathrm{hid}}}$ denote the input, recurrent, and readout weights, respectively. Here, $n_{\mathrm{hid}}=150$ for state estimation and $n_{\mathrm{hid}}=130$ for control, while $d_{\mathrm{in}}$ and $d_{\mathrm{out}}$ depend on the corresponding subnetwork. For state estimation, $\mathbf{p}[t]$ is the output of the preceding feedforward layer and for control, it is the external controller input.

\section{Embedded Benchmark Protocol and Deployment Footprint}
\label{app:embedded_benchmark}

  \paragraph{Benchmark scope.}
  We benchmarked matched ANN, dense SNN, and event-driven SNN implementations on the ESP32-S3. All three implementations execute the cascaded estimation and control networks using FP32 parameters. The dense and event-driven SNNs use identical model parameters and differ only in the implementation of spike-mediated synaptic propagation.

  \paragraph{Deployment footprint.}
  Model-array storage was obtained from the corresponding symbols in the linked ELF images. Resident RAM was measured on-device as the reduction in available internal heap immediately before and after network construction, initialization, and parameter loading. PSRAM was disabled for this measurement, so all dynamically allocated model weights, network states, and runtime buffers resided in the ESP32-S3 internal SRAM. The heap measurements include allocator alignment and memory-allocation overhead, but exclude FreeRTOS task stacks, IMU and serial buffers, and other firmware infrastructure.

  \begin{table}[t]
  \centering
  \scriptsize
  \caption{\textbf{Deployed model footprint on ESP32-S3.}
  Resident RAM denotes the dynamic internal heap allocated during model
  construction and loading. $1\,\mathrm{KiB}=1024$ bytes.}
  \label{tab:deployed_footprint}
  \vspace{1.0em}
  \begin{tabular}{@{}lrr@{}}
  \toprule
  \textbf{Implementation} &
  \textbf{Flash arrays} &
  \textbf{Resident RAM} \\
  \midrule
  ANN              & 251.13 KiB & 256.98 KiB \\
  Dense SNN        & 256.17 KiB & 273.66 KiB \\
  Event-driven SNN & 256.17 KiB & 274.57 KiB \\
  \bottomrule
  \end{tabular}
  \vspace{-1.0em}
  \end{table}

  \paragraph{Inference latency.}
  Inference latency was measured around the forward function only. IMU acquisition, input preparation, output serialization, and communication were excluded. Under the live-IMU workload used for Fig.~\ref{fig:esp32_latency_power}, the complete event-driven forward pass, including both the estimation and control networks, exhibited observed execution times of $610$--$750\,\mu\mathrm{s}$. Runtime model data were placed entirely in internal SRAM. Because event-driven execution depends on spike activity, $750\,\mu\mathrm{s}$ is the maximum latency observed under the tested workload rather than an analytically derived worst-case execution-time bound.

\paragraph{SIMD evaluation.}
A supplementary FP32 SIMD evaluation was conducted on the ESP32-S3 at
$240\,\mathrm{MHz}$ using ESP-IDF 5.5.4, with the same FP32 weights used
throughout the paper. The ANN, dense SNN, and event-driven SNN achieved
mean inference latencies of $648.722 \pm 0.199\,\mu$s,
$682.415 \pm 0.281\,\mu$s, and $557.560 \pm 51.160\,\mu$s, respectively,
corresponding to reductions of $38.7\%$, $37.6\%$, and $18.0\%$ relative
to their non-SIMD (scalar) implementations.

Dense FP32 projections use the ESP-DSP \texttt{dsps\_dotprod\_f32\_aes3()} kernel to optimize contiguous FP32 dot-product and multiply--accumulate operations. The buffers are 16-byte-aligned and their lengths are padded to multiples of four. For spike-mediated connections in the event-driven SNN, active presynaptic indices first select the corresponding weights. These weights are reorganized in memory so that the weights for four consecutive postsynaptic neurons are contiguous, enabling four-value loading with the ESP32-S3 \texttt{ee.ldf.128.ip} instruction and parallel accumulation. Sparse index traversal, indirect addressing, neuron-state updates, and scalar tails remain unvectorized. Because event-driven propagation already skips inactive neurons, SIMD covers a smaller fraction of the total computation and therefore provides a smaller additional speedup.

  \paragraph{Power measurement.}
  Board-level power was measured at the $5\,\mathrm{V}$ USB input of an
  ESP32-S3-DevKitC-1 using an inline FNIRSI FNB58 USB power meter. Wireless
  communication was disabled, and only the IMU was connected. Each condition
  was averaged over $60\,\mathrm{s}$. The inference-disabled baseline used
  otherwise identical firmware. Inference-induced power was calculated as
  \begin{equation}
  P_{\mathrm{inf}} = P_{\mathrm{total}}-P_{\mathrm{baseline}}.
  \label{eq:inference_power}
  \end{equation}

  The event-driven SNN therefore reduced baseline-subtracted inference power by $6\,\mathrm{mW}$, or approximately $18\%$, relative to the ANN. These values represent differences in board-level USB-input power and should not be interpreted as direct measurements of processor-core power alone.

\end{document}

%% file: acronym.tex
\acrodef{MAV}[MAV]{Micro Aerial Vehicle}

\acrodef{FWMAV}[FWMAV]{Flapping-Wing Micro Aerial Vehicle}

\acrodef{SNN}[SNN]{Spiking Neural Network}

\acrodef{ANN}[ANN]{Artificial Neural Network}

\acrodef{SWaP}[SWaP]{Size, Weight, and Power}

\acrodef{RL}[RL]{Reinforcement Learning}

\acrodef{IL}[IL]{Imitation Learning}

\acrodef{CPG}[CPG]{Central Pattern Generator}

\acrodef{IMU}[IMU]{Inertial Measurement Unit}

\acrodef{PWM}[PWM]{Pulse-Width Modulation}

\acrodef{PID}[PID]{Proportional-Integral-Derivative}

\acrodef{RLS}[RLS]{Recursive Least-Squares}

\acrodef{PPO}[PPO]{Proximal Policy Optimization}

\acrodef{UAV}[UAV]{Unmanned Aerial Vehicle}

\acrodef{LQR}[LQR]{Linear Quadratic Regulator}

\acrodef{MPC}[MPC]{Model Predictive Control}

\acrodef{6-DoF}[6-DoF]{six-degree-of-freedom}

\acrodef{LIF}[LIF]{Leaky Integrate-and-Fire}

\acrodef{NRP}[NRP]{Neurorobotics Platform}

\acrodef{RSTDP}[R-STDP]{Reward-modulated Spike-Timing-Dependent Plasticity}

\acrodef{CUBALIF}[CUBA-LIF]{Current-based Leaky Integrate-and-Fire}

\acrodef{BPTT}[BPTT]{Backpropagation Through Time}

\acrodef{RMSE}[RMSE]{Root Mean Squared Error}

\acrodef{DVS}[DVS]{Dynamic Vision Sensor}

\acrodef{FPGA}[FPGA]{Field-Programmable Gate Array}

\acrodef{ReLU}[ReLU]{Rectified Linear Unit}

\acrodef{SIMD}[SIMD]{Single Instruction, Multiple Data}

%% file: reference.bib
@article{nekoo2025survey,
  title={A review on flapping-wing robots: Recent progress and challenges},
  author={Rafee Nekoo, Saeed and Rashad, Ramy and De Wagter, Christophe and Fuller, Sawyer B and Croon, Guido de and Stramigioli, Stefano and Ollero, Anibal},
  journal={The International Journal of Robotics Research},
  pages={02783649251343638},
  year={2025},
  publisher={SAGE Publications Sage UK: London, England}
}

@article{hsiao2025rtmpc_il,
  title={Aerobatic maneuvers in insect-scale flapping-wing aerial robots via deep-learned robust tube model predictive control},
  author={Hsiao, Yi-Hsuan and Tagliabue, Andrea and Matteson, Owen and Kim, Suhan and Zhao, Tong and How, Jonathan P and Chen, YuFeng},
  journal={Science Advances},
  volume={11},
  number={49},
  pages={eaea8716},
  year={2025},
  publisher={American Association for the Advancement of Science}
}

@article{eshraghian2023training,
  title={Training spiking neural networks using lessons from deep learning},
  author={Eshraghian, Jason K and Ward, Max and Neftci, Emre O and Wang, Xinxin and Lenz, Gregor and Dwivedi, Girish and Bennamoun, Mohammed and Jeong, Doo Seok and Lu, Wei D},
  journal={Proceedings of the IEEE},
  volume={111},
  number={9},
  pages={1016--1054},
  year={2023},
  publisher={IEEE}
}

@article{bing2018survey,
  title={A survey of robotics control based on learning-inspired spiking neural networks},
  author={Bing, Zhenshan and Meschede, Claus and R{\"o}hrbein, Florian and Huang, Kai and Knoll, Alois C},
  journal={Frontiers in Neurorobotics},
  volume={12},
  pages={35},
  year={2018},
  publisher={Frontiers Media SA}
}

@article{oikonomou2025reinforcement,
  title={Reinforcement Learning with Spiking Neural Networks for Robotic Applications: A Survey},
  author={Oikonomou, Katerina Maria and Kansizoglou, Ioannis and Gasteratos, Antonios},
  journal={Authorea Preprints},
  year={2025},
  publisher={Authorea}
}

@inproceedings{clawson2016spiking,
  title={Spiking neural network ({SNN}) control of a flapping insect-scale robot},
  author={Clawson, Taylor S and Ferrari, Silvia and Fuller, Sawyer B and Wood, Robert J},
  booktitle={2016 IEEE 55th Conference on Decision and Control (CDC)},
  pages={3381--3388},
  year={2016},
  organization={IEEE}
}

@inproceedings{clawson2017adaptive,
  title={An adaptive spiking neural controller for flapping insect-scale robots},
  author={Clawson, Taylor S and Stewart, Terrence C and Eliasmith, Chris and Ferrari, Silvia},
  booktitle={2017 IEEE Symposium Series on Computational Intelligence (SSCI)},
  pages={1--7},
  year={2017},
  organization={IEEE}
}

@article{stroobants2025neuromorphic,
  title={Neuromorphic Attitude Estimation and Control},
  author={Stroobants, Stein and De Wagter, Christophe and De Croon, Guido CHE},
  journal={IEEE Robotics and Automation Letters},
  year={2025},
  publisher={IEEE}
}

@article{phan2019insect,
  title={Insect-inspired, tailless, hover-capable flapping-wing robots: Recent progress, challenges, and future directions},
  author={Phan, Hoang Vu and Park, Hoon Cheol},
  journal={Progress in Aerospace Sciences},
  volume={111},
  pages={100573},
  year={2019},
  publisher={Elsevier}
}

@article{hao2025insect,
  title={Insect-inspired passive mechanisms in hovering flapping wing micro air vehicles: A review},
  author={Hao, Jinjing and Wu, Jianghao},
  journal={Bioinspiration \& Biomimetics},
  year={2025}
}

@inproceedings{vanspiking,
  title={Spiking Neural Networks for High-Speed Continuous Quadcopter Control Using Proximal Policy Optimization},
  author={van Breukelen Castillo, MF and Ferede, R and Vos, RW and de Croon, C De Wagter GCHE},
  booktitle={16th International Micro Air Vehicle Conference and Competition},
  year={2025}
}

@article{stroobants2022neuromorphic,
  title={Neuromorphic computing for attitude estimation onboard quadrotors},
  author={Stroobants, Stein and Dupeyroux, Julien and de Croon, Guido CHE},
  journal={Neuromorphic Computing and Engineering},
  volume={2},
  number={3},
  pages={034005},
  year={2022},
  publisher={IOP Publishing}
}

@inproceedings{jiang2020target,
  title={Target tracking control of a wheel-less snake robot based on a supervised multi-layered {SNN}},
  author={Jiang, Zhuangyi and Otto, Richard and Bing, Zhenshan and Huang, Kai and Knoll, Alois},
  booktitle={2020 IEEE/RSJ International Conference on Intelligent Robots and Systems (IROS)},
  pages={7124--7130},
  year={2020},
  organization={IEEE}
}

@inproceedings{yang2025adaptive,
  title={Adaptive Wall-Following Control for Unmanned Ground Vehicles Using Spiking Neural Networks},
  author={Yang, Hengye and Chen, Yanxiao and Fan, Zexuan and Shao, Lin and Sun, Tao},
  booktitle={2025 IEEE/RSJ International Conference on Intelligent Robots and Systems (IROS)},
  pages={3786--3791},
  year={2025},
  organization={IEEE}
}

@inproceedings{stagsted2020towards,
  title={Towards neuromorphic control: A spiking neural network based {PID} controller for {UAV}},
  author={Stagsted, Rasmus and Vitale, Antonio and Binz, Jonas and Bonde Larsen, Leon and Sandamirskaya, Yulia and others},
  year={2020},
  organization={Robotics: Science and Systems (RSS)}
}

@inproceedings{bing2019end,
  title={End to end learning of a multi-layered {SNN} based on {R-STDP} for a target tracking snake-like robot},
  author={Bing, Zhenshan and Jiang, Zhuangyi and Cheng, Long and Cai, Caixia and Huang, Kai and Knoll, Alois},
  booktitle={2019 International Conference on Robotics and Automation (ICRA)},
  pages={9645--9651},
  year={2019},
  organization={IEEE}
}

@article{sharvit2025deep,
  title={A Deep Inverse-Mapping Model for a Flapping Robotic Wing},
  author={Sharvit, Hadar and Karl, Raz and Beatus, Tsevi},
  journal={arXiv preprint arXiv:2502.09378},
  year={2025}
}

@inproceedings{ross2011reduction,
  title={A reduction of imitation learning and structured prediction to no-regret online learning},
  author={Ross, St{\'e}phane and Gordon, Geoffrey and Bagnell, Drew},
  booktitle={Proceedings of the 14th International Conference on Artificial Intelligence and Statistics},
  pages={627--635},
  year={2011},
  organization={JMLR Workshop and Conference Proceedings}
}

@inproceedings{cai2025learning,
  title={Learning-based trajectory tracking for bird-inspired flapping-wing robots},
  author={Cai, Jiaze and Sangli, Vishnu and Kim, Mintae and Sreenath, Koushil},
  booktitle={2025 American Control Conference (ACC)},
  pages={430--437},
  year={2025},
  organization={IEEE}
}

@article{xiong2023lift,
  title={Lift enhancement of a butterfly-like flapping wing vehicle by reinforcement learning algorithm},
  author={Xiong, Min and Wei, Zhen and Yang, Yunjie and Chen, Qin and Liu, XiYan},
  journal={Bioinspiration \& Biomimetics},
  volume={18},
  number={4},
  pages={046010},
  year={2023},
  publisher={IOP Publishing}
}

@article{jafferis2019untethered,
  title={Untethered flight of an insect-sized flapping-wing microscale aerial vehicle},
  author={Jafferis, Noah T and Helbling, E Farrell and Karpelson, Michael and Wood, Robert J},
  journal={Nature},
  volume={570},
  number={7762},
  pages={491--495},
  year={2019},
  publisher={Nature Publishing Group UK London}
}

@article{de2009design,
  title={Design, aerodynamics, and vision-based control of the {DelFly}},
  author={De Croon, GCHE and De Clercq, KME and Ruijsink, Remes and Remes, Bart and De Wagter, Christophe},
  journal={International Journal of Micro Air Vehicles},
  volume={1},
  number={2},
  pages={71--97},
  year={2009},
  publisher={SAGE Publications Sage UK: London, England}
}

@article{phan2017design,
  title={Design and stable flight of a 21 g insect-like tailless flapping wing micro air vehicle with angular rates feedback control},
  author={Phan, Hoang Vu and Kang, Taesam and Park, Hoon Cheol},
  journal={Bioinspiration \& Biomimetics},
  volume={12},
  number={3},
  pages={036006},
  year={2017},
  publisher={IOP Publishing}
}

@inproceedings{zhang2017design,
  title={Design optimization and system integration of robotic hummingbird},
  author={Zhang, Jian and Fei, Fan and Tu, Zhan and Deng, Xinyan},
  booktitle={2017 IEEE International Conference on Robotics and Automation (ICRA)},
  pages={5422--5428},
  year={2017},
  organization={IEEE}
}

@article{gu202626,
  title={A 26-Gram Butterfly-Inspired Robot Achieving Autonomous Tailless Flight},
  author={Gu, Weibin and Feng, Chenrui and Liu, Lian and Yang, Chen and Jiao, Xingchi and Ding, Yuhe and Shi, Xiaofei and Gao, Chao and Rizzo, Alessandro and Zhou, Guyue},
  journal={arXiv preprint arXiv:2602.06811},
  year={2026}
}

@article{phan2024twist,
  title={A twist of the tail in turning maneuvers of bird-inspired drones},
  author={Phan, Hoang-Vu and Floreano, Dario},
  journal={Science Robotics},
  volume={9},
  number={96},
  pages={eado3890},
  year={2024},
  publisher={American Association for the Advancement of Science}
}

@article{ramezani2017biomimetic,
  title={A biomimetic robotic platform to study flight specializations of bats},
  author={Ramezani, Alireza and Chung, Soon-Jo and Hutchinson, Seth},
  journal={Science Robotics},
  volume={2},
  number={3},
  pages={eaal2505},
  year={2017},
  publisher={American Association for the Advancement of Science}
}

@inproceedings{huangdiffuseloco,
  title={{DiffuseLoco}: Real-Time Legged Locomotion Control with Diffusion from Offline Datasets},
  author={Huang, Xiaoyu and Chi, Yufeng and Wang, Ruofeng and Li, Zhongyu and Peng, Xue Bin and Shao, Sophia and Nikolic, Borivoje and Sreenath, Koushil},
  booktitle={8th Annual Conference on Robot Learning}
}
